\documentclass[runningheads]{llncs}

\usepackage{eccv}

\usepackage{eccvabbrv}

\usepackage{graphicx}
\usepackage{booktabs}

\usepackage[accsupp]{axessibility}  %

\usepackage{hyperref}

\usepackage{orcidlink}

\usepackage{soul}
\usepackage{multirow}
\usepackage{xcolor}
\usepackage{xspace}
\newcommand{\whitetxt}[1]{{\color{white}#1}\normalfont}
\usepackage{color}
\usepackage[dvipsnames]{xcolor}
\usepackage{subcaption}
\usepackage{epsfig}
\usepackage{graphicx,animate}
\usepackage{tabularx}
\usepackage{wrapfig}

\newbox\jsavebox

\definecolor{salmon}{RGB}{255, 117, 31}
\definecolor{teaser_orange}{RGB}{204, 78, 0}
\definecolor{teaserPurple}{RGB}{94, 23, 235}

\definecolor{R1color}{RGB}{0, 0, 255}      %
\definecolor{R2color}{RGB}{0, 176, 240}      %
\definecolor{R3color}{RGB}{0, 255, 0}       %

\newcommand{\methodname}{MV-Forcing}

\begin{document}

\title{\methodname{}: Long Multi-View Video Generation via 4D-Grounded Spatio-Temporal Self-Forcing} 

\author{
Gal Fiebelman\inst{1} \and
Hadar Averbuch-Elor\inst{2} \and
Sagie Benaim\inst{1}}

\institute{The Hebrew University of Jerusalem \and Cornell University}

\titlerunning{MV-Forcing}
\authorrunning{G.~Fiebelman et al.}
\maketitle
\vspace{-3mm}
\begin{center}
    \url{https://galfiebelman.github.io/mv-forcing/}
\end{center}

\begin{abstract}
Recent advances in video diffusion models have enabled either long single-view generation through temporal autoregression, or short multi-view synthesis through bidirectional attention. However, generating long, multi-view consistent videos of dynamic scenes remains unsolved. In this work, we present MV-Forcing, a framework that composes temporal and view-wise autoregression within a single diffusion model by introducing a 4D geometric bridge between sequentially generated views. Our key insight is that an autoregressive 3D reconstruction model naturally interfaces between autoregressively generated views. Given a completed source view, we reconstruct its 3D structure and render a geometric prior of the next target viewpoint, which the diffusion model refines into a high-quality video. To extend generation beyond the teacher's fixed temporal window, we introduce a joint denoising regime where both view slots are initialized from noise during training, enabling temporally unbounded generation. We distill the model via Distribution Matching Distillation with Spatio-Temporal Self-Forcing, closing the train-inference exposure bias gap for both temporal and view-sequential autoregression. Extensive experiments on both synthetic and real-world data demonstrate that MV-Forcing produces geometrically consistent multi-view videos of dynamic scenes at arbitrary lengths and viewpoint counts using a single few-step student model.

\end{abstract}


\begin{figure}[t]
    \centering
    \includegraphics[width=0.99\textwidth, trim=0cm 0.0cm 0cm 0cm,clip]{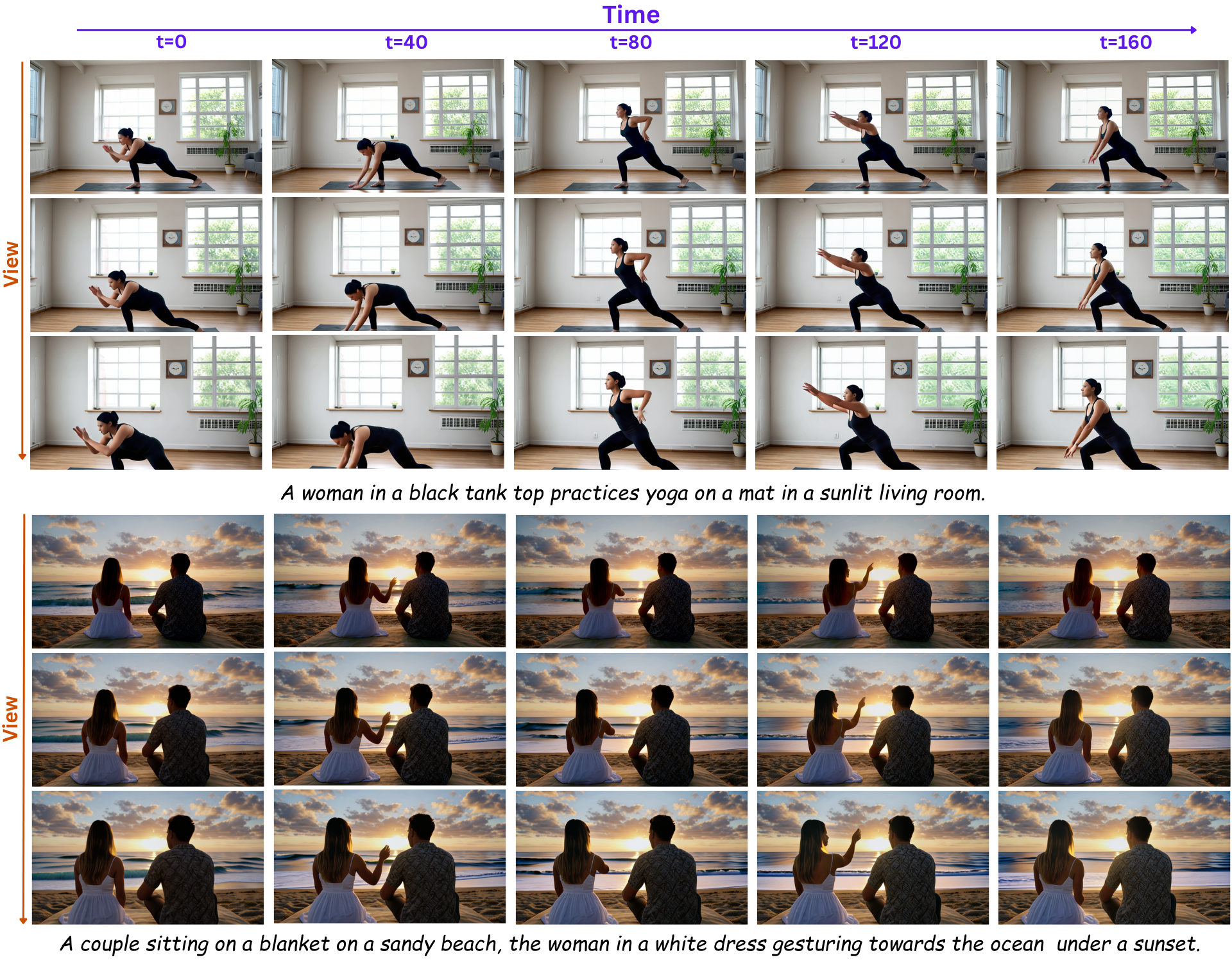}
    \vspace{-2mm}
    \caption{Given a text prompt and camera sequences, \methodname{} generates coherent video across an arbitrary number of viewpoints and unbounded temporal horizons. For each prompt, we show 3 \textcolor{teaser_orange}{views} at increasing camera displacements (rows) across 160 \textcolor{teaserPurple}{timesteps} (columns). Appearance, motion, and scene geometry remain consistent across all views and timesteps, demonstrating the effectiveness of our 4D-grounded geometric prior in maintaining cross-view consistency over long-horizon generation.}
    \vspace{-7mm}
\label{fig:teaser}
\end{figure}

\section{Introduction}
\label{sec:intro}

Generating a long, temporally coherent video from multiple viewpoints simultaneously is a long-standing challenge in computer vision. This task, long 
multi-view video generation, requires a model to synthesize the continuous temporal dynamics of a scene while maintaining strict 3D geometric consistency across arbitrary camera trajectories. Unlocking this capability is critical for a wide array of downstream applications, ranging from immersive virtual reality and advanced cinematic content creation to building dynamic, interactive simulations. However, jointly modeling the complex distribution of photorealistic video across both unbounded temporal horizons and an arbitrary number of viewpoints demands an intricate balance of spatial, temporal, and geometric reasoning.

Recent years have witnessed remarkable progress along two orthogonal directions in video synthesis. On one hand, continuous advances in autoregressive video diffusion have pushed the boundaries of temporal duration, enabling single-view generation over minute-long horizons by sequentially conditioning on previously generated context \cite{wang2025wan,huang2025self,liu2025rolling}. On the other hand, multi-view generation has seen exciting progress, with models now capable of synchronizing dynamic open-world scenes across multiple camera viewpoints \cite{bai2024syncammaster,kuang2024collaborative}. However, despite these parallel successes, unifying these paradigms to achieve long, multi-view generation remains a formidable bottleneck. Current multi-view approaches heavily rely on bidirectional attention across the entire time-view grid to enforce 3D consistency. While effective for short clips, this dense all-to-all attention scales quadratically, completely preventing streaming inference and severely restricting outputs to short, fixed temporal windows. Conversely, directly adapting temporal autoregressive models to the multi-view setting introduces severe exposure bias; without an explicit geometric anchor, sequential view generation quickly drifts, failing to maintain rigid spatial consistency. Consequently, generating videos that are simultaneously unbounded in time and geometrically consistent across arbitrary viewpoints remains an important, unsolved challenge.

In this work,  
we present \methodname{},  
a novel framework that uniquely composes temporal and view-wise autoregression within a single generative model, enabling the generation of long multi-view videos; see Figure \ref{fig:teaser}. Our core insight is that an autoregressive 4D reconstruction model (like CUT3R~\cite{wang2025continuous}) naturally interfaces between autoregressively generated views, serving as a continuous geometric bridge that circumvents the reliance on dense bidirectional attention. Given a completed source view, we reconstruct its 3D structure and render a geometric prior of the next target viewpoint. 
The generative model integrates this structural prior via a 3D convolution layer, refining it into a video alongside temporal context provided by the previously generated frames' KV cache.
By leveraging a recurrent reconstruction model (CUT3R), as each view is generated, it is incrementally integrated into a persistent latent state. This allows accumulating an increasingly complete 4D reconstruction of the scene, aligning reconstruction and generation within a unified autoregressive loop.

To extend generation beyond the teacher's fixed temporal window, we introduce a joint denoising regime where all 
view slots are initialized from noise during training, enabling temporally unbounded generation. 
Specifically, during training, we stochastically initialize both view slots from pure noise, which unifies first-view text-to-video generation and view-sequential conditioning within a single architecture.
Furthermore, to close the train/inference gap for both temporal and view-sequential autoregression, we distill our model via Distribution Matching Distillation (DMD) paired with Self-Forcing. 
By unrolling autoregressive generation along both axes and applying the DMD loss to the self-generated outputs, our Spatio-Temporal Self-Forcing mechanism effectively mitigates exposure bias, ensuring robust geometric and temporal coherence over long horizons.

To the best of our knowledge, we provide the first framework enabling long multi-view video generation. As such, we compare our method to baselines tackling either single-view long video generation or short multi-view generation. Specifically, we evaluate on both synthetic and real-world data, comparing against the bidirectional SynCamMaster teacher for short sequences, and state-of-the-art composed baselines (Self-Forcing and ReCamMaster) for long sequences. Quantitatively, we show that our framework maintains robust cross-view synchronization and strict camera accuracy even when scaling generation to arbitrary viewpoint counts and temporal lengths, effectively eliminating the severe geometric drift seen in prior approaches.

Our main contributions are: \textbf{(1).} 
We introduce \methodname{}, which is, to the best of our knowledge, the first generative framework to enable long multi-view video generation by uniquely composing temporal and view-wise autoregression. \textbf{(2).} We propose using an autoregressive 4D reconstruction model as a continuous geometric bridge between sequentially generated views, effectively bypassing the scaling limitations of dense bidirectional attention. \textbf{(3).} We introduce a joint denoising regime combined with Distribution Matching Distillation (DMD) and Spatio-Temporal Self-Forcing, enabling temporally unbounded, geometrically consistent synthesis closing the train-inference exposure bias gap. 

\section{Related Work}
\label{sec:related}

\noindent \textbf{Autoregressive Video Generation.}
While video diffusion models achieve remarkable quality~\cite{wang2025wan,yang2024cogvideox,polyak2024movie, kong2024hunyuanvideo}, most rely on bidirectional attention, preventing streaming and scaling. Autoregressive formulations alternatively generate frames sequentially.
Recent works combine autoregressive temporal structure with continuous diffusion to generate frame chunks. Teacher Forcing~\cite{hu2024acdit,jin2024pyramidal,gao2024ca2,zhang2025test} conditions on clean context, whereas Diffusion Forcing~\cite{chen2024diffusion,song2025history,gu2025long,magi1,chen2025skyreels} uses independently sampled noise levels per frame.
CausVid~\cite{yin2025causvid} introduces asymmetric distillation, training a causal student against a bidirectional teacher via DMD~\cite{yin2024improved,yin2024one}.
Self-Forcing~\cite{huang2025self} identifies that both paradigms still suffer from exposure bias and addresses this by unrolling autoregressive generation during training. Several concurrent works build on this paradigm~\cite{liu2025rolling,yang2025longlive,cui2025self} to push long-horizon quality further. All of these methods operate in the single-view regime. Our work extends Self-Forcing to the spatio-temporal domain, composing temporal autoregression with view-sequential generation grounded by a 4D geometric prior.

\smallskip
\noindent \textbf{Multi-View Video Generation.}
Object-centric methods such as SV4D~\cite{xie2024sv4d}, SV4D~2.0~\cite{yao2025sv4d}, and CAT4D~\cite{wu2025cat4d} generate dense orbital views of dynamic objects but are restricted to single-object scenes and fixed camera configurations. For open-world scenes, SynCamMaster~\cite{bai2024syncammaster} introduces a plug-and-play multi-view synchronization module for a pretrained text-to-video DiT, enabling multi-camera generation with 6-DoF camera control. CVD~\cite{kuang2024collaborative} adds epipolar attention across views to enforce 3D consistency. ReCamMaster~\cite{bai2025recammaster} and related approaches~\cite{van2024generative,jeong2025reangle} condition on a reference video to re-render it from novel trajectories, but are video-to-video methods and do not generate multi-view content from text. All these methods rely on bidirectional attention over both time and views, restricting generation to short, fixed-length clips and a fixed number of views. Our method extends Self-Forcing to the multi-view setting, enabling generation over unbounded temporal horizons and an arbitrary number of viewpoints.

\smallskip
\noindent \textbf{Reconstruction for Generation.}
A growing body of work leverages 3D reconstruction as a structural prior for generation by reconstructing geometry from available observations, rendering it from target viewpoints, and conditioning a generative model on these renders to fill in appearance details~\cite{fridman2023scenescape,ren2025gen3c,yu2024viewcrafter,yu2025trajectorycrafter,li2025vmem,gu2025diffusion,bian2025gs,fiebelman2026let}. Underlying many of these methods are feed-forward reconstruction models such as DUSt3R~\cite{wang2024dust3r}, MASt3R~\cite{leroy2024grounding}, VGGT~\cite{wang2025vggt}, and CUT3R~\cite{wang2025continuous}, which predict dense pointmaps and camera poses without per-scene optimization. Among these, CUT3R is uniquely suited to sequential generation due to its recurrent architecture, it maintains a persistent latent state updated incrementally with each new image, and supports querying from arbitrary virtual cameras. Our work leverages CUT3R as a geometric bridge between sequentially generated views, aligning reconstruction and generation within a unified autoregressive loop.

\begin{figure*}[t!]
    \centering
    \includegraphics[width=0.99\textwidth]{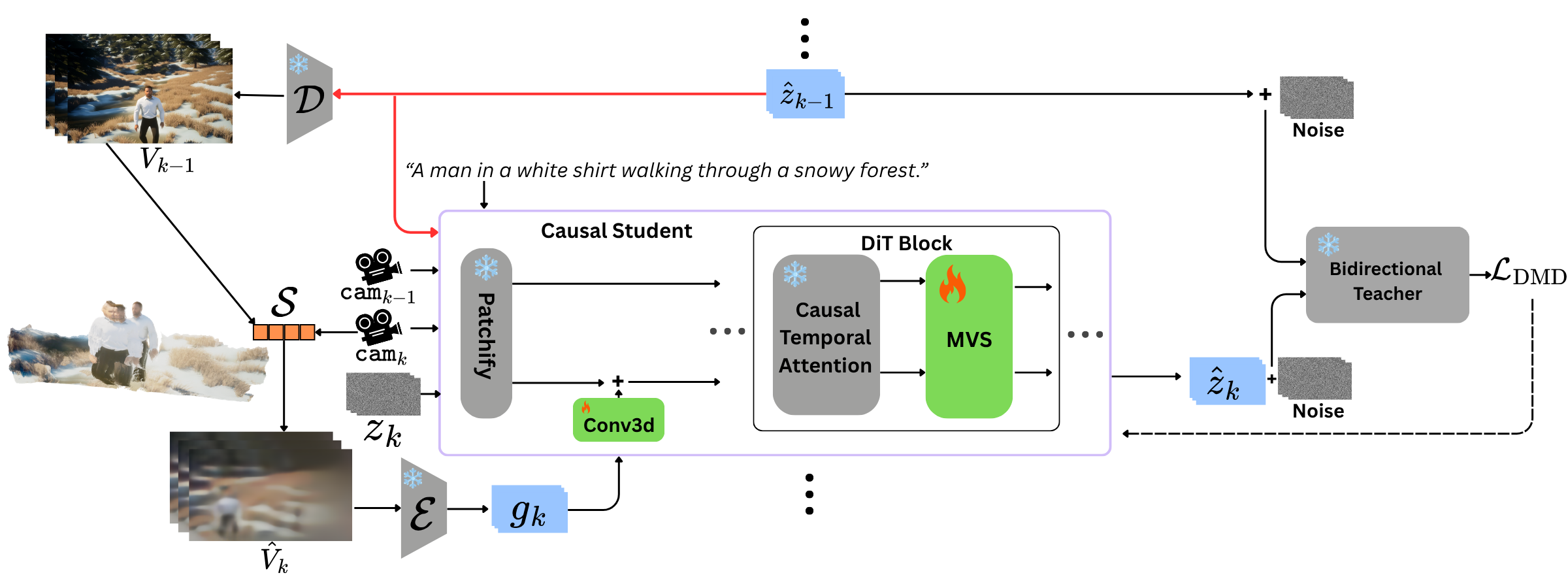}
        \vspace{-2mm}
    \caption{\textbf{Overview of MV-Forcing.} Given a text prompt, camera sequences $\texttt{cam}_{k-1}$ and $\texttt{cam}_k$, and a previously generated view $\hat{z}_{k-1}$, the causal student generates the next view $\hat{z}_k$. The decoded previous view $V_{k-1} = \mathcal{D}(\hat{z}_{k-1})$ is integrated into CUT3R's persistent state $\mathcal{S}$, which is queried using $\texttt{cam}_k$ to produce a geometric rendering $\hat{V}_k$. This rendering is encoded and projected via a $\texttt{Conv3d}$ layer into a conditioning tensor $g_k$ that is added to the patchified noisy latent $z_k$. Both $\hat{z}_{k-1}$ (clean) and $z_k$ (noisy) are forwarded through DiT blocks with causal temporal attention and MVS cross-view attention. The student's output $\hat{z}_k$ and the preceding $\hat{z}_{k-1}$ are re-noised and scored by the bidirectional SynCamMaster teacher via $\mathcal{L}_{\text{DMD}}$. The \textcolor{red}{red arrow} illustrates the view-sequential unrolling, $\hat{z}_k$ is decoded and integrated into CUT3R's state, and fed back as the conditioning view for the next generation step, mirroring the inference-time autoregressive loop. The 3D reconstruction visualized alongside $\mathcal{S}$ illustrates the scene geometry accumulated in the state from all previously generated views.}
\vspace{-2mm}
\label{fig:overview}
\end{figure*}

\section{Method}
\label{sec:method}
Given an input text prompt and a set of $N$ camera sequences, each defining a desired viewpoint, our goal is to generate a long, multi-view consistent video that scales to an arbitrary number of viewpoints and extends over unbounded temporal horizons. As illustrated in Fig.~\ref{fig:overview}, we achieve this by composing temporal and view-sequential autoregression within a unified self-forcing framework, using a feed-forward dynamic 3D reconstruction model as a geometric bridge between the two axes. We begin by reviewing the multi-view video generation and self-forcing paradigms that form the basis of our approach (Sec.~\ref{sec:prelim}). We then introduce spatio-temporal self-forcing, which extends the self-forcing paradigm from the temporal dimension to the view dimension (Sec.~\ref{sec:st_self_forcing}). Next, we describe the 4D-grounded geometric prior that bridges temporal and view-sequential generation with an explicit geometric signal (Sec.~\ref{sec:geo_prior}). Finally, we describe how these components compose at inference to produce long multi-view videos (Sec.~\ref{sec:inference})  and how the framework generalizes to real-world data (Sec.~\ref{sec:real_world}).

\subsection{Preliminaries}
\label{sec:prelim}

\noindent \textbf{Multi-View Video Generation.}
SynCamMaster~\cite{bai2024syncammaster} extends a pretrained text-to-video diffusion transformer (DiT)~\cite{wang2025wan} to multi-view generation by introducing a plug-and-play multi-view synchronization~(MVS) module. Given a pair of videos $V_1, V_2$ from two viewpoints, a 3D VAE encoder $\mathcal{E}$ compresses each into a latent representation $z_j = \mathcal{E}(V_j)$, which is then converted into a sequence of tokens via patchification:
\begin{equation}
    {x}_1 = \texttt{patchify}({z}_1), \quad {x}_2 = \texttt{patchify}({z}_2),
\end{equation}
where $x_1, x_2 \in \mathbb{R}^{T \times S \times D}$, with $T$ the number of latent frames, $S$ the spatial tokens per frame, and $D$ the token dimension.  Each transformer block applies bidirectional temporal self-attention across all frames within a single view. An MVS module is additionally inserted into each transformer block. Given the intermediate features $F_1^i, F_2^i \in \mathbb{R}^{S \times D}$ at frame $i$ within a transformer block, the module first adds camera embeddings produced by a camera encoder $\mathcal{E}_c$ that maps the 6-DoF extrinsic parameters into the token space, and then applies cross-view self-attention, where tokens from both views at the same frame attend to each other:
\begin{equation}
    \overline{{F}}_1^i, \overline{{F}}_2^i = \texttt{MVS}({F}_1^i + \mathcal{E}_c(\texttt{cam}_1), \; {F}_2^i + \mathcal{E}_c(\texttt{cam}_2)).
\label{eq:mvs}
\end{equation}
The output is projected back to the feature domain by a zero-initialized linear layer and added as a residual to the original features. Only the MVS parameters are trained, while the base DiT weights remain frozen. In SynCamMaster, both temporal and cross-view attention are bidirectional, and generation follows a many-step denoising schedule that requires all frames and views to be denoised jointly within a fixed-length window.

\smallskip
\noindent \textbf{Self-Forcing.}
A common strategy to accelerate video diffusion is to distill the model into a few-step generator via Distribution Matching Distillation~(DMD)~\cite{yin2024improved,yin2024one}. CausVid~\cite{yin2025causvid} applies asymmetric DMD to distill a bidirectional video DiT into a causal few-step student. Self-Forcing~\cite{huang2025self} identifies that the student still suffers from exposure bias, as it trains on ground-truth context but must condition on its own imperfect outputs at inference. It addresses this by unrolling autoregressive generation during training and applying the DMD loss over the self-generated output, closing the train-inference gap and enabling streaming generation of arbitrarily long single-view videos.

\subsection{Spatio-Temporal Self-Forcing}
\label{sec:st_self_forcing}
While Self-Forcing~\cite{huang2025self} enables temporal autoregression for long single-view videos, extending this scalability across arbitrary viewpoints requires view-wise autoregression.
We propose \emph{spatio-temporal self-forcing}, which extends self-forcing to the view dimension by introducing view-sequential autoregressive unrolling during training. 
Following Self-Forcing~\cite{huang2025self}, we distill SynCamMaster into a causal few-step student $G_\phi$ via asymmetric DMD~\cite{yin2024improved}. The student inherits the teacher's MVS module (Eq.~\ref{eq:mvs}), but replaces the bidirectional temporal attention with causal attention under a blockwise mask:
\begin{equation}
    M_{i,j} =
    \begin{cases}
    1, & \text{if } \lfloor \frac{j}{K} \rfloor \leq \lfloor \frac{i}{K} \rfloor, \\
    0, & \text{otherwise},
    \end{cases}
\end{equation}
where $i$ and $j$ index latent frames in the sequence and $K$ is the temporal block size in latent frames. Under this mask, tokens at frame $i$ attend only to frames within the same or earlier temporal blocks.

The student is trained via asymmetric DMD, where the frozen bidirectional SynCamMaster teacher serves as the data score function $s_{\text{data}}$ and a trainable copy $s_{\text{gen}}$ tracks the student's output distribution. At each training step, the student generates an output $\hat{z}_0 = G_\phi(\{z^i_{\tau^i}\}, \{\tau^i\})$ via its few-step schedule, where $z^i_{\tau^i}$ is the $i$-th temporal block corrupted at independently sampled noise level $\tau^i$. Noise is re-injected at a shared level $\tau$ to obtain $\hat{z}_\tau$, and the student is updated via:
\begin{equation}
    \nabla_\phi \mathcal{L}_{\text{DMD}} \approx -\mathbb{E}_{\tau}\left[\left(s_{\text{data}}(\hat{z}_\tau, \tau) - s_{\text{gen}}(\hat{z}_\tau, \tau)\right) \frac{\partial \hat{z}_\tau}{\partial \phi}\right].
\label{eq:dmd}
\end{equation}
Following CausVid~\cite{yin2025causvid}, we first initialize the student by training on a small set of ODE solution pairs generated by the bidirectional teacher before applying the DMD objective. The student's temporal layers are initialized from a pretrained Self-Forcing model and kept frozen, while the MVS layers are initialized from SynCamMaster and finetuned during training. Additional training details are provided in Sec.~\ref{sec:imp_details}.

\smallskip
\noindent \textbf{View-Sequential Generation.}
Given $N$ target viewpoints, we generate views $0, 1, \ldots, N{-}1$ sequentially. To produce view $k$, we denoise its latent starting from pure noise $z_k^{\tau_Q} \sim \mathcal{N}(0, I)$ through a $Q$-step schedule $\tau_Q > \cdots > \tau_1 > \tau_0 = 0$, conditioned on the fully denoised preceding view and the text prompt $P$. At each denoising step $q = Q, \ldots, 1$, the student takes the noisy latent $z_k^{\tau_q}$ along with the clean preceding latent $z_{k-1}$, text prompt $P$, and camera parameters $\texttt{cam}_{k}, \texttt{cam}_{k-1}$, predicts a clean estimate, and re-noises it to the next level:
\begin{equation}
    z_k^{\tau_{q-1}} = \Psi\!\left(G_\phi(z_k^{\tau_q},\, \tau_q,\, z_{k-1},\, P,\, \texttt{cam}_k, \texttt{cam}_{k-1}),\; \tau_{q-1}\right),
\label{eq:view_denoise}
\end{equation}
where $G_\phi$ is the student network and $\Psi(\cdot,\, \tau_{q-1})$ re-injects noise at level $\tau_{q-1}$. The final output $z_k = z_k^{\tau_0}$ then serves as clean conditioning for view $k{+}1$ via the cross-view attention in Eq.~\ref{eq:mvs}, naturally extending to arbitrary-length view chains. Since each view only attends to the single preceding view, the process is fully autoregressive across viewpoints. Note that this clean-noisy asymmetry across views is absent from the SynCamMaster teacher, which jointly denoises both views at the same noise level. The student learns this clean cross-view conditioning entirely through distillation.

\smallskip
\noindent \textbf{View-Sequential Unrolling.}
This view-sequential generation introduces a view-level analogue of exposure bias. During DMD training, the cross-view attention conditions on ground-truth latents, but at inference, view $k$ is conditioned on the model's previously generated $z_{k-1}$. We close this gap by unrolling autoregressive generation along the view dimension during training. Starting from the ground-truth first view $z_0^{\text{gt}}$, the student sequentially generates views $\hat{z}_1, \hat{z}_2, \ldots$, conditioned on its previously generated output, via its few-step schedule (Eq.~\ref{eq:view_denoise}):
\vspace{-0.5cm}
\begin{align}
    \hat{z}_{k-1} &= G_\phi(\epsilon_{k-1},\, \{\tau_q\},\, \hat{z}_{k-2},\, P,\, \texttt{cam}_{k-1}, \texttt{cam}_{k-2}), \\
    \hat{z}_k &= G_\phi(\epsilon_k,\, \{\tau_q\},\, \hat{z}_{k-1},\, P,\, \texttt{cam}_k, \texttt{cam}_{k-1}),
\end{align}
where $\hat{z}_0 = z_0^{\text{gt}}$, $\epsilon_k \sim \mathcal{N}(0, I)$, $\{\tau_q\}$ denotes the few-step denoising schedule, and we abbreviate the full denoising chain (Eq.~\ref{eq:view_denoise}) for brevity. Each generated view beyond the first is conditioned on the student's own imperfect output rather than ground-truth, mirroring the inference-time regime and forcing the student to learn robustness to its own errors. The bidirectional SynCamMaster teacher then scores each consecutive generated pair $(\hat{z}_{k-1}, \hat{z}_k)$ via the DMD loss (Eq.~\ref{eq:dmd}). Since the teacher evaluates view pairs independently, the distillation objective naturally decomposes over both the temporal and view axes, jointly mitigating the exposure-bias gap along both dimensions. In practice, due to GPU memory constraints, we unroll from $z_0^{\text{gt}}$ to generate $\hat{z}_1$ and $\hat{z}_2$.

\smallskip
\noindent \textbf{Joint View Denoising.}
As our goal is text to multi-view video generation, the model must also generate the first view from text prompt $P$ alone, without any cross-view conditioning. To this end, during training we set both views to start from noise with probability $p$, training the model to generate each view independently from text. When only view $k$ starts from noise (probability $1 - p$), the model trains the view-sequential conditioning path described above. This training regime unifies first-view generation and view-sequential generation within a single architecture. At inference, the model generates $z_0$ from text via the joint denoising path with both view latents initialized from noise, and then generates each subsequent view autoregressively conditioned on the preceding one.

\subsection{4D-Grounded Geometric Prior}
\label{sec:geo_prior}

As views are generated sequentially over long temporal horizons and across multiple viewpoints, geometric consistency degrades as each view is conditioned only on the single preceding view through the cross-view attention in Eq.~\ref{eq:mvs}, and errors in 3D structure accumulate along the view chain. To mitigate this, we introduce an explicit geometric conditioning signal that bridges temporal and view-sequential generation, grounding each new viewpoint in a persistent 3D reconstruction of the scene built from all previously generated views and timesteps.

\smallskip
\noindent \textbf{Recurrent 3D Reconstruction.}
Our geometric prior requires a 3D reconstruction model that satisfies two properties: it must operate autoregressively since views and temporal blocks are generated sequentially, and it must maintain a persistent state that can be queried from arbitrary virtual cameras, enabling rendering of geometric priors for viewpoints and timesteps not yet generated.
CUT3R~\cite{wang2025continuous} is a feed-forward dynamic 3D reconstruction model that meets both requirements. It operates recurrently over a stream of images, maintaining a persistent latent state $\mathcal{S}$ that encodes the accumulated 4D structure of the scene. Given a new image $I$, the state is updated:
\begin{equation}
    \mathcal{S}' = \texttt{CUT3R\_update}(\mathcal{S}, I),
\end{equation}
where $\mathcal{S}$ is the state prior to observing $I$ and $\mathcal{S}'$ is the updated state. This state can also be queried from arbitrary virtual cameras without updating it. Given a target camera parameterized as a raymap $\mathcal{R}$~\cite{zhang2024cameras,gao2024cat3d,wu2025cat4d}, CUT3R decodes a rendered image and a per-pixel confidence map:
\begin{equation}
    \hat{I}, \; \hat{C} = \texttt{CUT3R\_query}(\mathcal{S}, \mathcal{R}),
\label{eq:cut3r_query}
\end{equation}
where $\hat{I}$ is the RGB rendering of the scene from the queried viewpoint, $\hat{C}$ is a confidence map indicating reconstruction certainty at each pixel, and $\mathcal{R}$ is a 6-channel image encoding ray origins and directions at each pixel. 

\smallskip
\noindent \textbf{Geometric Conditioning.}
To condition the generation of view $k$, we first decode all preceding views to pixel space, $V_j = \mathcal{D}(z_j)$ for $j = 0, \ldots, k{-}1$, where $\mathcal{D}$ is the 3D VAE decoder, and integrate them into CUT3R's state via sequential updates. We then query the accumulated state from view $k$'s camera sequence, for each frame at time $t$, we construct a raymap $\mathcal{R}_k^t$ from the camera extrinsics and intrinsics and obtain a rendering $\hat{I}_k^t$ and confidence map $\hat{C}_k^t$ via Eq.~\ref{eq:cut3r_query}. The sequence of rendered frames $\hat{V}_k = \{\hat{I}_k^1, \ldots, \hat{I}_k^T\}$ is encoded into video latents $\mathcal{E}(\hat{V}_k)$ and concatenated channel-wise with the confidence maps to form a conditioning tensor:
\begin{equation}
    g_k = \text{Concat}\!\left(\mathcal{E}(\hat{V}_k), \; \hat{C}_k\right) \in \mathbb{R}^{T \times D' \times H \times W},
\label{eq:geo_cond}
\end{equation}
where $D' = C_{\text{latent}} + 1$ combines the VAE latent channels with the single-channel confidence map. A 3D convolution layer projects the conditioning tensor into the token space and adds it to the patch embeddings via a residual connection:
\begin{equation}
    \tilde{x}_k = x_k + \texttt{Conv3d}(g_k),
\end{equation}
where $x_k = \texttt{patchify}(z_k)$ are the patch embeddings of view $k$. Following ControlNet~\cite{zhang2023adding}, the Conv3d layer is zero-initialized, ensuring the geometric prior has no effect when training begins and allowing the model to gradually learn to incorporate it. Together with the clean cross-view conditioning described in Sec.~\ref{sec:st_self_forcing}, the geometric prior constitutes a second conditioning pathway that the student acquires entirely through distillation, as neither is present in the original teacher model.

\smallskip
\noindent \textbf{State Accumulation.}
As each temporal block of a view is generated, its latent $z_k$ is decoded to pixel space $V_k = \mathcal{D}(z_k)$ and fed into CUT3R, updating $\mathcal{S}$ incrementally along both time and views. When generating block $(k, t)$, the state already incorporates all decoded frames from earlier views $0, \ldots, k{-}1$ and preceding temporal blocks $(k, <t)$ of the current view. Additional details of the accumulated geometric prior are provided in Sec.~\ref{sec:geo_prior_details}.

\subsection{Long Multi-View Video Generation}
\label{sec:inference}

The autoregressive structure along both axes, and the accumulated geometric prior, yields a key property at inference, generation can advance along \emph{either} axis. At any block $(k, t)$, the model can proceed temporally to $(k, t+1)$ or across views to $(k{+}1, t)$, since both transitions require only the KV cache (for temporal context) and a clean preceding view with its geometric prior (for cross-view context). This allows arbitrary traversal orders over the time$\times$view grid, rather than requiring one axis to be fully completed before advancing the other.

Concretely, the first block $(0, 0)$ is generated from the text prompt alone using joint view denoising (Sec.~\ref{sec:st_self_forcing}). For each subsequent block $(k, t)$, the generated latent is decoded to pixel space $V_k = \mathcal{D}(z_k)$ and integrated into CUT3R's state $\mathcal{S}$. When advancing temporally within a view, the KV cache provides temporal context for the next block. When advancing to a new view $k$, the accumulated state is queried from view $k$'s cameras to produce the geometric conditioning tensor $g_k$ (Eq.~\ref{eq:geo_cond}), which along with the clean preceding latent $z_{k-1}$ provides cross-view conditioning through the MVS module (Eq.~\ref{eq:mvs}). In both cases, the decoded frames are fed back into CUT3R, progressively enriching the state along both axes. By the time generation reaches view $k{+}1$, the accumulated state captures the full 4D structure observed so far, providing increasingly rich geometric context for each successive viewpoint.

\subsection{Generalization to Real-World Data}
\label{sec:real_world}

As the DMD objective (Eq.~\ref{eq:dmd}) distills the teacher's distribution through the data score function $s_{\text{data}}$, the student inherits the domain of its teacher, which in our case is the synthetic SynCamMaster. Extending the framework to real-world content requires replacing $s_{\text{data}}$ with a teacher whose score function captures a real-world distribution. As no such multi-view teacher exists, we adopt ReCamMaster~\cite{bai2025recammaster}, a video-to-video model that re-renders a source video from a novel camera trajectory, and finetune the student for $N_{\text{ft}}$ iterations on real-world videos to bridge the distribution shift. As ReCamMaster conditions on an existing video rather than generating from text, it cannot supervise the joint denoising path (Sec.~\ref{sec:st_self_forcing}), and we set $p{=}0$ during this stage. At inference, we compose Self-Forcing~\cite{huang2025self} for first-view generation with our finetuned student for subsequent views. Additional details are provided in Sec.~\ref{sec:real_world_details}.

\section{Experiments}
\label{sec:experiments}

We evaluate \methodname{} on the task of long multi-view video generation. We present comparisons against baselines in Sec.~\ref{sec:comparisons}, ablate our design choices and analyze scaling behavior along both the view and temporal axes in Sec.~\ref{sec:ablations}. Finally, we discuss limitations in Sec.~\ref{sec:limitations}. 

\smallskip
\noindent \textbf{Evaluation Setup.}
We train and evaluate our model in two settings. For the synthetic setting, we use the SynCamVideo Dataset~\cite{bai2024syncammaster}, which contains 3,400 synthetic multi-view video scenes, each captured from 10 synchronized cameras, and hold out 100 scenes for evaluation. For the real-world setting, we follow the extension described in Sec.~\ref{sec:real_world}. We finetune our model on real-world videos from the Mixkit subset of the Open-Sora Dataset~\cite{opensoradata}, which contains diverse open-domain video content, and hold out 100 videos for evaluation. All metrics are averaged across the evaluation set. Additional evaluation details are provided in Sec.~\ref{sec:eval_details}.

\smallskip
\noindent \textbf{Evaluation Metrics.}
Following SynCamMaster~\cite{bai2024syncammaster}, we evaluate our method along three axes, visual quality, camera accuracy, and cross-view synchronization. For visual quality, we report  Fr\'{e}chet Inception Distance~(FID)~\cite{heusel2017gans}, Fr\'{e}chet Video Distance~(FVD)~\cite{unterthiner2019fvd}, CLIP-T, which measures average CLIP~\cite{radford2021learning} similarity between
each frame and its text prompt and CLIP-F, which measures average CLIP similarity between consecutive frames. For camera accuracy, we use GIM~\cite{shen2024gim} to estimate per-frame camera poses from
dense correspondences and report the average rotation and translation error against the ground truth, denoted as RotErr and TransErr, respectively~\cite{he2024cameractrl}. For cross-view synchronization, we report Mat.~Pix, which measures the number of matching pixels with confidence above a threshold from the computed dense correspondences between view pairs from GIM, CLIP-V~\cite{kuang2024collaborative}, which measures average CLIP similarity between views at the same timestep, and FVD-V~\cite{xie2024sv4d}, which measures FVD of stacked frames across views at each timestep. 

\subsection{Comparison to Baselines}
\label{sec:comparisons}

\noindent \textbf{Baselines.}
No existing method addresses long multi-view video generation. We therefore construct baselines by composing single-view, multi-view, and video-to-video methods. SynCamMaster~\cite{bai2024syncammaster} is the bidirectional many-step teacher, generating 2 views in a fixed 81-frame window, serving as a strong short sequence baseline, but cannot scale beyond this setting. SF+ReCamMaster uses Self-Forcing~\cite{huang2025self} to generate an arbitrarily long first view, then applies ReCamMaster~\cite{bai2025recammaster} to re-render additional views in independent 81-frame segments. SF+ReCamMaster+SF uses Self-Forcing to generate an 81-frame first view, then applies ReCamMaster and uses Self-Forcing to temporally extend each re-rendered view beyond its first segment. For the synthetic evaluation, we additionally finetune Self-Forcing on SynCamVideo, replacing its real-world teacher with the synthetic SynCamMaster, denoted SF(ft), to ensure a fair in-domain comparison. Additional baseline details are provided in Sec.~\ref{sec:baseline_details}.

\begin{table}[t]
    \centering
    \caption{\textbf{Short Sequence Evaluation.} All methods generate 2 views at 81 frames. We compare MV-Forcing against the bidirectional SynCamMaster~\cite{bai2024syncammaster} teacher.}
    \label{tab:short_comparisons}
    \vspace{-3mm}
    \resizebox{\linewidth}{!}{%
    \begin{tabular}{lcccc|cc|ccc}
    \toprule
    & \multicolumn{4}{c|}{Visual Quality} & \multicolumn{2}{c|}{Camera Accuracy} & \multicolumn{3}{c}{View Synchronization}\\
    \cmidrule(r){2-10}
    Method & FID $\downarrow$ & FVD $\downarrow$ & CLIP-T $\uparrow$ & CLIP-F $\uparrow$ & RotErr $\downarrow$ & TransErr $\downarrow$ & Mat.~Pix.(K) $\uparrow$ & FVD-V $\downarrow$ & CLIP-V $\uparrow$\\
    \midrule
    SynCamMaster          & \textbf{166.57} & \textbf{1451.44} & \textbf{30.02} & \textbf{99.32} & 3.83 & 8.83 & 236.72 & 1697.37 & 90.13 \\
    MV-Forcing (Ours)     & 167.90 & 1468.84 & 29.67 & 99.21 & \textbf{3.64} & \textbf{8.26} & \textbf{251.13} & \textbf{1691.05} & \textbf{91.81} \\
    \bottomrule
    \end{tabular}%
    }
    \vspace{-3mm}
\end{table}

\begin{table}[t]
    \centering
\caption{\textbf{Long Sequence Evaluation.} All methods generate 3 views at 162 frames. SF+ReCamMaster generates each view independently from a Self-Forcing~\cite{huang2025self} generated first view then re-renders with ReCamMaster~\cite{bai2025recammaster}. SF+ReCamMaster+SF additionally extends each view temporally using Self-Forcing after the initial ReCamMaster re-rendering. We evaluate in two settings, real-world (Real), where all methods operate on real-world data, and synthetic (Synth.), where we replace Self-Forcing with SF(ft) finetuned on SynCamVideo for a fair in-domain comparison.}    \label{tab:long_comparisons}
    \vspace{-3mm}
    \resizebox{\linewidth}{!}{%
    \begin{tabular}{llcccc|cc|ccc}
    \toprule
    & & \multicolumn{4}{c|}{Visual Quality} & \multicolumn{2}{c|}{Camera Accuracy} & \multicolumn{3}{c}{View Synchronization}\\
    \cmidrule(r){3-11}
    & Method & FID $\downarrow$ & FVD $\downarrow$ & CLIP-T $\uparrow$ & CLIP-F $\uparrow$ & RotErr $\downarrow$ & TransErr $\downarrow$ & Mat.~Pix.(K) $\uparrow$ & FVD-V $\downarrow$ & CLIP-V $\uparrow$\\
    \midrule
    \multirow{3}{*}{\rotatebox{90}{\small Real}}
    & SF+ReCamMaster & 157.57 & 1397.42 & 32.27 & 98.22 & 4.74 & 10.12 & 146.81 & 1759.82 & 87.26 \\
    & SF+ReCamMaster+SF & 156.78 & 1363.23 & 32.48 & 98.67 & 4.89 & 10.61 & 127.48 & 1771.34 & 86.61 \\
    & MV-Forcing (Ours) & \textbf{153.27} & \textbf{1309.68} & \textbf{32.74} & \textbf{99.03} & \textbf{3.88} & \textbf{8.78} & \textbf{239.37} & \textbf{1554.23} & \textbf{90.83} \\
    \midrule
    \multirow{3}{*}{\rotatebox{90}{\small Synth.}}
    & SF(ft)+ReCamMaster & 192.34 & 1576.83 & 29.07 & 98.03 & 4.32 & 10.27 & 143.77 & 1956.38 & 87.19 \\
    & SF(ft)+ReCamMaster+SF & 191.26 & 1592.78 & 29.26 & 98.52 & 4.67 & 10.73 & 121.29 & 1987.43 & 86.27 \\
    & MV-Forcing (Ours) & \textbf{186.73} & \textbf{1560.54} & \textbf{29.54} & \textbf{99.17} & \textbf{3.72} & \textbf{8.41} & \textbf{243.63} & \textbf{1729.35} & \textbf{89.88} \\
    \bottomrule
    \end{tabular}%
    }
    \vspace{-5mm}
\end{table}

\begin{figure*}[t!]
    \centering
    \includegraphics[width=\textwidth, trim={2.7cm 8.5cm 4.0cm 3.2cm}, clip]{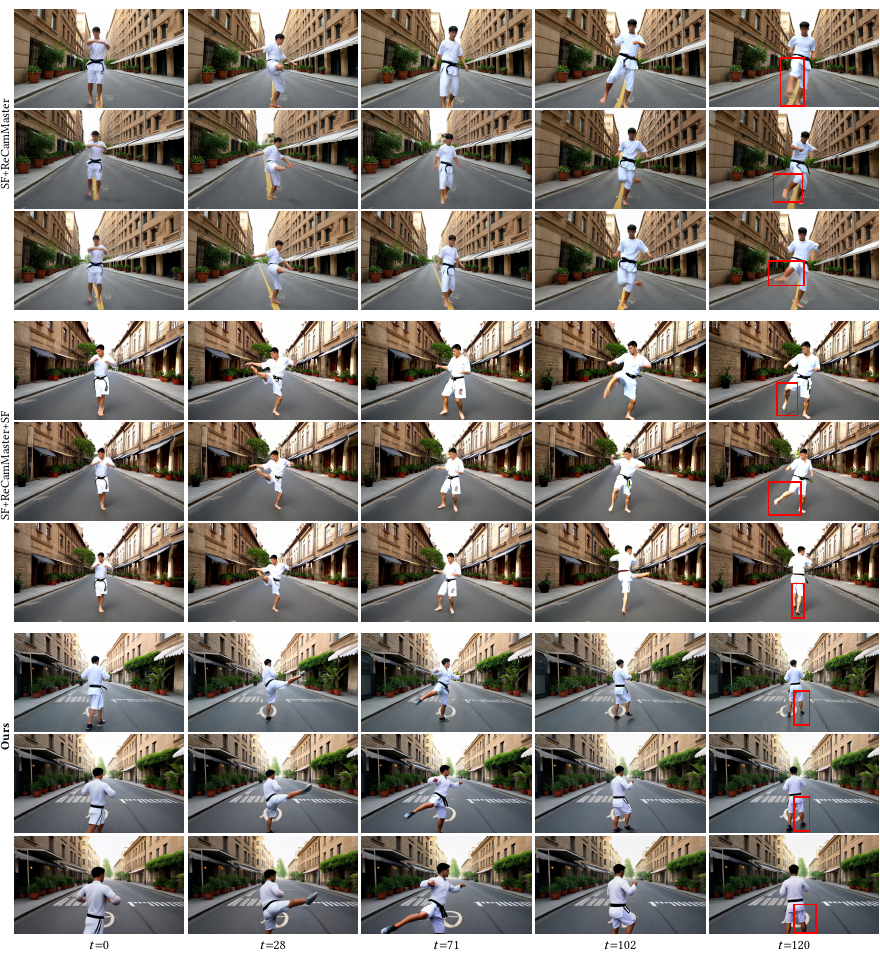}
    \vspace{-3mm}
    \caption{\textbf{Qualitative Comparison on Long Sequences.} We generate multiple views across 162 frames for the prompt \emph{``A young man wearing white clothes practicing karate kicks on an urban street.''} For each method, we show 3 views (rows) at 5 timesteps (columns). SF+ReCamMaster degrades when operating beyond its training window, producing visible artifacts in the third view and beyond 81 frames. SF+ReCamMaster+SF recovers temporal continuity but lacks cross-view consistency, with scene structure and appearance varying significantly between viewpoints. Our method maintains consistency across both views and time. As highlighted by the \textcolor{red}{red} boxes (shown on the last timestep for visibility), the baselines produce inconsistent leg positions across views at the same timestep, while our method preserves a coherent pose across all viewpoints.}
\label{fig:long_comparisons}
\end{figure*}

\smallskip
\noindent \textbf{Comparisons.}
We evaluate under two settings. Tab.~\ref{tab:short_comparisons} compares against SynCamMaster at 2 views and 81 frames. Tab.~\ref{tab:long_comparisons} extends the evaluation to 3 views and 162 frames, beyond SynCamMaster's capacity. We compare against both SF+ReCamMaster and SF+ReCamMaster+SF in the real-world setting, and against their SF(ft) variants in the synthetic setting.

In the short sequence setting, MV-Forcing outperforms the bidirectional teacher on all camera accuracy and view synchronization metrics, demonstrating that explicit geometric grounding through the accumulated CUT3R prior can outperform joint bidirectional denoising even on short sequences. Visual quality metrics remain slightly lower, reflecting the expected cost of distilling a many-step bidirectional model into a few-step causal student. In the long sequence setting (Tab.~\ref{tab:long_comparisons}), MV-Forcing outperforms all baselines across all metrics in both evaluation settings. SF+ReCamMaster generates each view as a sequence of independent 81-frame segments with no shared context between them, leading to visible degradation beyond the first segment. SF+ReCamMaster+SF recovers temporal continuity by extending each view autoregressively, but lacks any cross-view conditioning mechanism, and consistency between viewpoints degrades as views are generated independently. The consistent results across both real-world and synthetic settings indicate that the performance gap is architectural rather than driven by domain alignment. These results demonstrate that view-sequential conditioning with a persistent geometric prior provides the 3D grounding necessary for cross-view consistency, while temporal autoregression ensures coherent generation beyond any fixed temporal window.

A qualitative comparison is shown in Fig.~\ref{fig:long_comparisons}. SF+ReCamMaster produces plausible results within its first temporal window but degrades at later timesteps due to independent per-segment generation. SF+ReCamMaster+SF maintains temporal continuity but exhibits cross-view drift, with scene structure and appearance varying between viewpoints. In contrast, our method preserves geometric consistency across all views and timesteps. Qualitative comparisons for the short setting and synthetic long setting are provided in Sec.~\ref{sec:supp_comp}.

\subsection{Ablation Study}
\label{sec:ablations}

\begin{table}[t]
    \centering
    \caption{\textbf{Quantitative Ablation Results.}  All variants are evaluated at 3 views and 162 frames. We ablate view-sequential unrolling (w/o View Unrolling), the geometric prior (w/o CUT3R), state accumulation (w/o Accumulation), and CUT3R's latent state query (Manual Rendering).}
    \label{tab:component_ablations}
    \vspace{-3mm}
    \resizebox{\linewidth}{!}{%
    \begin{tabular}{lcccc|cc|ccc}
    \toprule
    & \multicolumn{4}{c|}{Visual Quality} & \multicolumn{2}{c|}{Camera Accuracy} & \multicolumn{3}{c}{View Synchronization}\\
    \cmidrule(r){2-10}
    Method & FID $\downarrow$ & FVD $\downarrow$ & CLIP-T $\uparrow$ & CLIP-F $\uparrow$ & RotErr $\downarrow$ & TransErr $\downarrow$ & Mat.~Pix.(K) $\uparrow$ & FVD-V $\downarrow$ & CLIP-V $\uparrow$\\
    \midrule
    w/o View Unrolling  & 212.89 & 1791.40 & 27.74 & 97.66 & 4.72 & 10.45 & 159.73 & 2318.94 & 86.41 \\
    w/o CUT3R & 199.10 & 1613.21 & 28.87 & 98.21 & 4.56 & 9.83 & 173.75 & 2109.82 & 87.12 \\
    w/o Accumulation & 193.66 & 1576.84 & 29.26 & 98.78 & 3.97 & 9.04 & 229.63 & 1984.35 & 88.27 \\
    Manual Rendering & 191.14 & 1571.44 & 29.11 & 99.10 & 3.93 & 8.93 & 232.71 & 1963.42 & 88.36 \\
    Ours & \textbf{186.73} & \textbf{1560.54} & \textbf{29.54} & \textbf{99.17} & \textbf{3.72} & \textbf{8.41} & \textbf{243.63} & \textbf{1729.35} & \textbf{89.88} \\
    \bottomrule
    \end{tabular}%
    }
    \vspace{-3mm}

\end{table}

We ablate the key components of our framework: (1)~\emph{w/o Accumulation} conditions each view only on the single preceding view's geometry; (2)~\emph{Manual Rendering} replaces CUT3R's learned raymap query with classical forward-warping via z-buffer splatting; (3)~\emph{w/o CUT3R} removes the geometric prior entirely; and (4)~\emph{w/o View Unrolling} trains with standard DMD along the view axis without spatio-temporal self-forcing unrolling. 

Tab.~\ref{tab:component_ablations} reports results at 3 views and 162 frames. Removing view-sequential unrolling yields the largest degradation, as the train-inference distribution mismatch propagates along the view chain. Removing CUT3R produces the second-largest drop, validating the geometric prior as critical for cross-view consistency. Removing accumulation and replacing CUT3R's query with classical warping each degrade metrics by a smaller but consistent margin. Qualitative comparisons of the ablation variants are provided in Sec.~\ref{sec:supp_comp}.

\begin{table}[t]
    \centering
   \caption{\textbf{View scaling.}  All variants are evaluated at a fixed length of 81 frames. We increase the number of generated views from 2 to 5.}
    \label{tab:view_ablations}
        \vspace{-3mm}

    \resizebox{\linewidth}{!}{%
    \begin{tabular}{lcccc|cc|ccc}
    \toprule
     & \multicolumn{4}{c|}{Visual Quality} & \multicolumn{2}{c|}{Camera Accuracy} & \multicolumn{3}{c}{View Synchronization}\\
    \cmidrule(r){2-10}
    Views & FID $\downarrow$ & FVD $\downarrow$ & CLIP-T $\uparrow$ & CLIP-F $\uparrow$ & RotErr $\downarrow$ & TransErr $\downarrow$ & Mat.~Pix.(K) $\uparrow$ & FVD-V $\downarrow$ & CLIP-V $\uparrow$\\
    \midrule
    \whitetxt{x}2 & \textbf{167.9} & \textbf{1468.84} & \textbf{29.67} & \textbf{99.21} & \textbf{3.64} & \textbf{8.26} & \textbf{251.13} & \textbf{1691.05} & \textbf{91.81} \\
    \whitetxt{x}3 & 168.33 & 1468.92 & 29.58 & 99.13 & 3.65 & 8.29 & 251.08 & 1691.86 & 91.77 \\
    \whitetxt{x}4 & 168.57 & 1469.03 & 29.55 & 99.06 & 3.65 & 8.31 & 251.02 & 1691.94 & 91.75 \\
    \whitetxt{x}5 & 169.13 & 1469.12 & 29.43 & 98.92 & 3.67 & 8.35 & 250.95 & 1692.07 & 91.73 \\
    \bottomrule
    \end{tabular}%
    }
    \vspace{-3mm}
\end{table}

\begin{table}[t]
    \centering
    \caption{\textbf{Temporal scaling.} All variants are evaluated at a fixed 2 views. We increase the temporal length from 81 to 648 frames.} 
    \label{tab:num_frames_ablations}
        \vspace{-3mm}

    \resizebox{\linewidth}{!}{%
    \begin{tabular}{lcccc|cc|ccc}
    \toprule
     & \multicolumn{4}{c|}{Visual Quality} & \multicolumn{2}{c|}{Camera Accuracy} & \multicolumn{3}{c}{View Synchronization}\\
    \cmidrule(r){2-10}
    Length & FID $\downarrow$ & FVD $\downarrow$ & CLIP-T $\uparrow$ & CLIP-F $\uparrow$ & RotErr $\downarrow$ & TransErr $\downarrow$ & Mat.~Pix.(K) $\uparrow$ & FVD-V $\downarrow$ & CLIP-V $\uparrow$\\
    \midrule
    \whitetxt{xi}81 & \textbf{167.90} & \textbf{1468.84} & \textbf{29.67} & \textbf{99.21} & \textbf{3.64} & \textbf{8.26} & \textbf{251.13} & \textbf{1691.05} & \textbf{91.81} \\
    \whitetxt{x}162 & 169.34 & 1529.52 & 29.59 & 99.19 & 3.66 & \textbf{8.26} & 249.82 & 1693.26 & 91.64 \\
    \whitetxt{x}324 & 170.36 & 1675.26 & 29.54 & 97.82 & 3.67 & 8.27 & 249.72 & 1694.84 & 91.57 \\
    \whitetxt{x}648 & 172.91 & 1859.82 & 29.52 & 96.23 & 3.67 & 8.29 & 249.14 & 1694.98 & 91.28 \\
    \bottomrule
    \end{tabular}%
    }
    \vspace{-5mm}
\end{table}

\smallskip
\noindent \textbf{Scaling Analysis.}
We additionally evaluate how generation quality evolves as the number of views and temporal length are independently extended.

Tab.~\ref{tab:view_ablations} reports metrics as the number of views
increases from 2 to 5 at a fixed temporal length of 81 frames.
All metrics show only minor degradation, remaining stable up to 5 views. We attribute this robustness to the accumulated geometric prior, grounding each new view in the full 4D structure. Tab.~\ref{tab:num_frames_ablations} extends temporal length from 81 to 648 frames at 2 views. Cross-view metrics remain nearly constant across an 8$\times$ increase in duration, while CLIP-F shows the most notable decline, reflecting the compounding of small temporal errors inherent to autoregressive temporal generation. Qualitative comparisons for both scaling axes are provided in Sec.~\ref{sec:supp_comp}.

\subsection{Limitations}
\label{sec:limitations}

While \methodname{} enables long multi-view video generation, several limitations remain. First, our model is primarily trained 
on the synthetic SynCamVideo dataset. Although we demonstrate 
generalization to real-world data with minimal finetuning 
(Sec.~\ref{sec:real_world}), training on large-scale real 
multi-view video data could further improve diversity and realism. Second, while Self-Forcing significantly reduces the train-inference exposure bias for temporal autoregression, quality still degrades over long horizons. Recent works~\cite{liu2025rolling,yang2025longlive,cui2025self} propose complementary strategies to mitigate this. These works are complementary to our framework and could be integrated to improve temporal stability. Third, as the SynCamMaster teacher generates only two views at a time, the DMD supervision never directly supervises consistency across more than two simultaneous views. Extending the training window to longer view tuples could further improve quality at high viewpoint counts. Despite this, Tab.~\ref{tab:view_ablations} shows stable performance up to 5 views at inference, as the model chains pairwise transitions autoregressively. We provide a further failure case analysis in Sec.~\ref{sec:failure_cases}, examining scenarios such as extreme camera displacements, extreme motion, and generation backbone quality.
\section{Conclusion}
\label{sec:conclusion}

We presented \methodname{}, a framework for long multi-view video generation that composes temporal and view-sequential autoregression within a single generative model. Our key insight is that an online dynamic 3D reconstruction model can serve as a geometric bridge between sequentially generated views, accumulating a persistent state that grounds each new viewpoint in the geometry of previously generated content. By extending Self-Forcing to the view axis, our approach mitigates the exposure bias gap along both axes, enabling generation over unbounded temporal horizons and arbitrary viewpoint counts with a single few-step model. We believe this takes a step toward spatially consistent, multi-view video generation that captures the full 4D structure of dynamic scenes.

\paragraph{Ethics Statement.} Our method builds on publicly available open-source models and datasets. Our contribution is long multi-view video generation, a capability that does not introduce new generative content beyond what existing open-source text-to-video models already produce.

\paragraph{Acknowledgements.} This research was supported by The Israel Science Foundation (grant No. 2416/25) and EuroHPC JU (Application ID EHPC-DEV-2025D08-098).

\bibliographystyle{splncs04}
\bibliography{main}

\clearpage
\appendix
\noindent {\LARGE\textbf{Appendix}}

\section{Interactive Visualizations}
\label{sec:interactive}

We provide interactive video visualizations at \href{https://galfiebelman.github.io/mv-forcing/supp/index.html}{https://galfiebelman.github.io/mv-forcing/supp/index.html}, including multi-view generation results across diverse scenes, real-world generalization examples, comparisons against baselines, component ablations, and view and temporal scaling results.

\section{Additional Details}
\label{sec:supp_details}

\subsection{Implementation Details}
\label{sec:imp_details}

\medskip \noindent \textbf{Architecture.}
Our student model inherits the Wan2.1-T2V-1.3B~\cite{wang2025wan}  (available at \href{https://huggingface.co/Wan-AI/Wan2.1-T2V-1.3B}{https://huggingface.co/Wan-AI/Wan2.1-T2V-1.3B}) backbone used by SynCamMaster~\cite{bai2024syncammaster} (available at \href{https://huggingface.co/KlingTeam/SynCamMaster-Wan2.1/blob/main/step20000.ckpt}{https://huggingface.co/KlingTeam/SynCamMaster-Wan2.1}). Bidirectional temporal attention is replaced with causal attention under the blockwise mask described in Sec.~\ref{sec:prelim}, with a temporal block size of $K = 3$ latent frames. The MVS module from SynCamMaster is retained in every transformer block. The temporal layers are initialized from a pretrained Self-Forcing~\cite{huang2025self} model (available at \href{https://huggingface.co/gdhe17/Self-Forcing/blob/main/checkpoints/self_forcing_dmd.pt}{https://huggingface.co/gdhe17/Self-Forcing}) and kept frozen, while the MVS layers are initialized from SynCamMaster and finetuned. The Conv3d geometric conditioning layer projects the CUT3R~\cite{wang2025continuous} (available at \href{https://drive.google.com/file/d/1Asz-ZB3FfpzZYwunhQvNPZEUA8XUNAYD/view?usp=drive_link}{https://github.com/CUT3R/CUT3R}) render and confidence map into the token dimension and is zero-initialized following ControlNet~\cite{zhang2023adding}.

\medskip \noindent \textbf{Training.}
Training consists of two stages, both performed on 16 NVIDIA A100 65GB GPUs using Adam ($\beta_1 = 0, \beta_2 = 0.999$) with batch size 1 per GPU (effective batch size 16).
In the first stage (ODE initialization), we generate 1{,}000 ODE solution pairs from the bidirectional SynCamMaster teacher and train the student for 3{,}000 iterations with a learning rate of $2 \times 10^{-6}$. This stage takes approximately 16 hours.
In the second stage (DMD distillation with spatio-temporal self-forcing), we train for 1{,}600 iterations with the asymmetric DMD loss using generator and critic learning rates of $2 \times 10^{-6}$ and $4 \times 10^{-7}$ respectively. The joint denoising probability is set to $p = 0.3$. This stage takes approximately 26 hours.
CUT3R~\cite{wang2025continuous} is kept frozen throughout both stages.

\subsection{Real-World Extension Details}
\label{sec:real_world_details}

As described in Sec.~\ref{sec:real_world}, we adopt ReCamMaster~\cite{bai2025recammaster} as the real-world teacher. The official ReCamMaster (available at 
\href{https://huggingface.co/KlingTeam/ReCamMaster-Wan2.1}{\url{https://huggingface.co/KlingTeam/ReCamMaster-Wan2.1}}) model assumes that the generated video shares its first frame with the reference video, whereas in our multi-view setting the target viewpoint begins from a completely different camera with no overlapping first frame. To adapt to this non-overlapping setting, we finetune ReCamMaster for 2{,}000 iterations on the SynCamVideo dataset with arbitrary viewpoint pairs. The same finetuned model is used for the baseline comparisons reported in Sec.~\ref{sec:baseline_details}. We set the joint denoising probability $p{=}0$ and train only the view-sequential conditioning path, applying the DMD loss (Eq.~\ref{eq:dmd}) with the finetuned ReCamMaster serving as the data score $s_{\text{data}}$. For training data, we use real-world videos from the Mixkit subset of the Open-Sora Dataset~\cite{opensoradata}, which serve as source latents for both the student and the ReCamMaster teacher. We finetune the student for 500 iterations on this data.

\subsection{Geometric Prior Details}
\label{sec:geo_prior_details}

\medskip \noindent \textbf{Accumulation Order.}
At inference, as each temporal block of a view is generated, it is decoded to pixel space via the 3D VAE decoder $\mathcal{D}$ and fed into CUT3R to update its persistent state $\mathcal{S}$. When generating view $k$ at temporal block $t$, the state already incorporates all frames from views $0, \ldots, k{-}1$ and from view $k$ at temporal blocks $0, \ldots, t{-}1$.
When advancing to view $k{+}1$, the accumulated state is queried via raymaps for all frames of view $k{+}1$'s camera sequence, producing the RGB rendering and confidence map used to construct the conditioning tensor $g_{k+1}$ (Eq.~\ref{eq:geo_cond}).

\medskip \noindent \textbf{Confidence Maps.}
CUT3R outputs a per-pixel confidence map $\hat{C} \in [0, 1]$ indicating reconstruction certainty. Regions visible from multiple previously generated views receive high confidence, while occluded or unobserved regions receive low confidence. The confidence map is concatenated channel-wise with the VAE-encoded CUT3R render (Eq.~\ref{eq:geo_cond}), allowing the Conv3d layer to learn to weight the geometric prior based on its reliability at each spatial location.

\begin{figure*}[t!]
    \centering
    \includegraphics[width=\textwidth, trim={3.0cm 22.4cm 4.5cm 3.2cm}, clip]{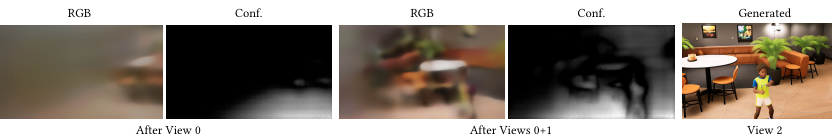}
    \caption{\textbf{CUT3R Geometric Prior Visualization.} We generate a 3-view video for the prompt \emph{``A child dances in a restaurant''} and query view 2's camera at the same timestep under two accumulation stages. For each stage, we show the rendered RGB image from CUT3R's~\cite{wang2025continuous} accumulated state and the corresponding per-pixel confidence map, where brighter values indicate higher reconstruction certainty. After only view 0, the RGB render is largely uninformative and confidence is near zero, as view 0 has minimal overlap with view 2. After integrating views 0 and 1, the rendered prior captures recognizable scene structure with substantially higher confidence. The rightmost column shows the final generated view 2, produced by the student conditioned on this accumulated prior.}
    \label{fig:cut3r_prior}
\end{figure*}

\medskip \noindent \textbf{Visualization.}
Fig.~\ref{fig:cut3r_prior} visualizes the geometric prior at different stages of accumulation. We generate a 3-view video for the prompt \emph{``A child dances in a restaurant''} and query view 2's camera at the same timestep under two accumulation stages. After only view 0, the rendered RGB is largely uninformative and the confidence map is near zero, as view 0 has small overlap with view 2. After integrating view 1, the prior captures recognizable scene structure and confidence increases significantly in previously occluded regions. This progressive enrichment provides increasingly informative conditioning as more views are generated.

\subsection{Runtime Analysis}
\label{sec:runtime}

We report wall-clock inference times and peak GPU memory on a single NVIDIA A100 65GB GPU. For a single view at 81 frames (4-step denoising), the student forward pass takes approximately 20 seconds. The CUT3R update and query for the geometric prior adds approximately 30 seconds per view transition. For reference, SynCamMaster requires approximately 6 minutes for 2 views at 81 frames due to its 50-step bidirectional denoising, and ReCamMaster similarly requires approximately 6 minutes per 81-frame chunk. Our 4-step causal student achieves a ${\sim}5\times$ speedup for the short-sequence setting while additionally supporting unbounded temporal and view scaling. Tab.~\ref{tab:runtime} reports runtime and peak VRAM across configurations. Runtime scales approximately linearly with view count and temporal length, while peak VRAM remains constant at 23~GB across all configurations.

\begin{table}[t]
    \centering
    \caption{\textbf{Inference scaling.} Wall-clock time and peak GPU memory on a single NVIDIA A100 65GB GPU across different view and temporal configurations.}
    \label{tab:runtime}
    \vspace{-3mm}
    \begin{tabular}{lcc}
    \toprule
    Configuration & Time & Peak VRAM \\
    \midrule
    2 views $\times$ 81 frames   & $\sim$70s   & 23 GB \\
    3 views $\times$ 162 frames  & $\sim$4 min & 23 GB \\
    5 views $\times$ 81 frames   & $\sim$4 min & 23 GB \\
    2 views $\times$ 648 frames  & $\sim$10 min & 23 GB \\
    5 views $\times$ 648 frames  & $\sim$30 min & 23 GB \\
    \bottomrule
    \end{tabular}
    \vspace{-3mm}
\end{table}

\subsection{Evaluation Details}
\label{sec:eval_details}

\medskip \noindent \textbf{Evaluation Protocol.}
We evaluate on 100 held-out scenes across three camera configurations (azimuthal rotation, elevation change, distance variation). All metrics are averaged across configurations and scenes. The same protocol and metrics are used for both the synthetic (SynCamVideo) and real-world (Mixkit) evaluation settings. Visual quality metrics (FID~\cite{heusel2017gans}, FVD~\cite{unterthiner2019fvd}, CLIP-T, CLIP-F) follow standard protocols using CLIP ViT-L/14~\cite{radford2021learning} for the CLIP-based metrics.

\medskip \noindent \textbf{Camera Accuracy Metrics.}
We use GIM~\cite{shen2024gim} to compute dense correspondences between consecutive view pairs at every 4th pixel frame. From 5{,}000 sampled sparse matches, we estimate the essential matrix via RANSAC and recover the relative pose. RotErr and TransErr~\cite{he2024cameractrl} report the average geodesic rotation error and angular translation error (in degrees) against the ground-truth relative poses.

\medskip \noindent \textbf{Cross-View Synchronization Metrics.}
Mat.~Pix reports the number of matching pixels (in thousands) with GIM~\cite{shen2024gim} confidence above 0.01 between consecutive view pairs, normalized to the original image resolution. CLIP-V measures the average CLIP~\cite{radford2021learning} cosine similarity between all view pairs at each timestep. FVD-V~\cite{xie2024sv4d} computes FVD on cross-view clips formed by stacking frames from all views at each timestep, padded to 10 frames by repetition when fewer views are available.

\subsection{Baseline Details}
\label{sec:baseline_details}

Below we provide all details needed to reproduce the baseline comparisons.

\medskip \noindent \textbf{SynCamMaster.}
We use the official SynCamMaster~\cite{bai2024syncammaster} model as provided by the authors (\href{https://github.com/KlingAIResearch/SynCamMaster}{https://github.com/KlingAIResearch/SynCamMaster}). The model is built on Wan2.1-T2V-1.3B with the default 50-step denoising schedule and classifier-free guidance scale of 5.0. This model generates 2 views jointly in a fixed 81-frame window using bidirectional temporal and cross-view attention.

\medskip \noindent \textbf{SF+ReCamMaster.}
We use Self-Forcing~\cite{huang2025self} with the code provided by the authors (\href{https://github.com/guandeh17/Self-Forcing}{https://github.com/guandeh17/Self-Forcing}) to generate the first view at arbitrary length using the pretrained DMD student with KV cache. For additional views, we use ReCamMaster~\cite{bai2025recammaster} with the code provided by the authors (\href{https://github.com/KlingAIResearch/ReCamMaster}{https://github.com/KlingAIResearch/ReCamMaster}) to re-render the first view from the target camera trajectory. We use the same finetuned ReCamMaster described in Sec.~\ref{sec:real_world_details}, adapted to the non-overlapping first-frame setting. Since ReCamMaster operates within a fixed 81-frame window, long sequences are generated as independent per-chunk segments with no shared context between them.

\medskip \noindent \textbf{SF+ReCamMaster+SF.}
This extends the above by using Self-Forcing to temporally continue each additional view beyond the first ReCamMaster chunk. ReCamMaster generates the initial 81-frame segment, and Self-Forcing extends it autoregressively using the ReCamMaster output as context.

\medskip \noindent \textbf{SF(ft).}
To ensure a fair in-domain comparison on the synthetic SynCamVideo evaluation set, we finetune Self-Forcing on SynCamVideo, replacing its real-world teacher with the synthetic SynCamMaster. We score only the generated first view $V_0$ and finetune for 1{,}000 iterations to bridge the domain gap. We denote the resulting model SF(ft) and use it to replace the Self-Forcing component in both baselines for the synthetic evaluation reported in Tab.~\ref{tab:long_comparisons}.

\section{Failure Case Analysis}
\label{sec:failure_cases}

\begin{figure}[t!]
  \centering
  \includegraphics[trim={7.75cm 13.0cm 7.9cm 13.75cm},clip, width=\textwidth]{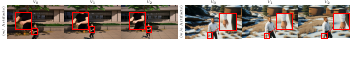}\\[-1.2cm]
  \includegraphics[trim={7.75cm 13.0cm 7.9cm 13.7cm},clip, width=\textwidth]{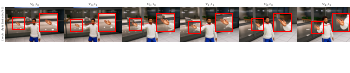}\\[-1.17cm]
  \includegraphics[trim={7.75cm 13.3cm 7.9cm 13.6cm},clip, width=\textwidth]{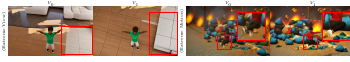}
  \vspace{-7mm}
\caption{\textbf{Failure Case Analysis.} \textbf{Top:} Generation backbone quality. When the first view $V_0$ contains artifacts such as malformed hands (left, red boxes), these are faithfully propagated to subsequent views through the autoregressive formulation. When $V_0$ is artifact-free (right), subsequent views remain clean. \textbf{Middle:} The artifact-free case extends across both views and timesteps, with consistent hand geometry maintained across all views and temporal steps. \textbf{Bottom left:} Extreme camera displacement. At the steeper viewing angle, sofa geometry is displaced onto the floor region. The geometric prior has depth ambiguity in regions unobserved from the source view, and cross-view attention cannot establish correct spatial correspondences. \textbf{Bottom right:} Extreme motion (e.g., kicking while jumping) causes CUT3R to reconstruct incorrect limb geometry, resulting in an inconsistent number of visible legs across views. In both cases, the model could in principle recover by relying on the previous view $\hat{z}_{k-1}$ rather than the geometric prior, but these scenarios are highly out-of-distribution, making this difficult.}
\label{fig:failure_cases}
\vspace{-3mm}
\end{figure}

We identify three categories of failure cases, illustrated in Fig.~\ref{fig:failure_cases}.

\medskip \noindent \textbf{Generation Backbone Quality.}
Artifacts produced in the first view $V_0$, which is generated from text alone without geometric conditioning, are propagated to subsequent views through the autoregressive formulation. These artifacts, such as malformed hands, also occur in general text-to-video models and in our SynCamMaster teacher. When $V_0$ does not contain such artifacts, subsequent views remain clean across both views and timesteps (Fig.~\ref{fig:failure_cases}, top and middle). Conversely, when $V_0$ contains artifacts, these are faithfully propagated (Fig.~\ref{fig:failure_cases}, top left). A stronger generation backbone or teacher model would mitigate this.

\medskip \noindent \textbf{Extreme Camera Displacement.}
When the target viewpoint differs significantly from the source, the geometric prior has depth ambiguity in regions unobserved from previous views, and cross-view attention cannot establish correct spatial correspondences. As shown in Fig.~\ref{fig:failure_cases} (bottom left), at the steeper viewing angle, sofa geometry is displaced onto the floor region.

\medskip \noindent \textbf{Extreme Motion.}
Fast or complex motion (e.g., kicking while jumping) can cause CUT3R to reconstruct incorrect limb geometry. The corrupted geometric prior provides unreliable spatial anchors for the target view, resulting in an inconsistent number of visible legs across views (Fig.~\ref{fig:failure_cases}, bottom right). In both extreme cases, the model could in principle recover by relying on the previous view $\hat{z}_{k-1}$ rather than the geometric prior, but these scenarios are highly out-of-distribution, making this difficult.

\section{Additional Comparisons and Ablations}
\label{sec:supp_comp}

We provide additional qualitative comparisons and ablations 
below.

\begin{figure*}[t!]
    \centering
    \includegraphics[width=\textwidth, trim={2.7cm 6.4cm 4.2cm 3.2cm}, clip]{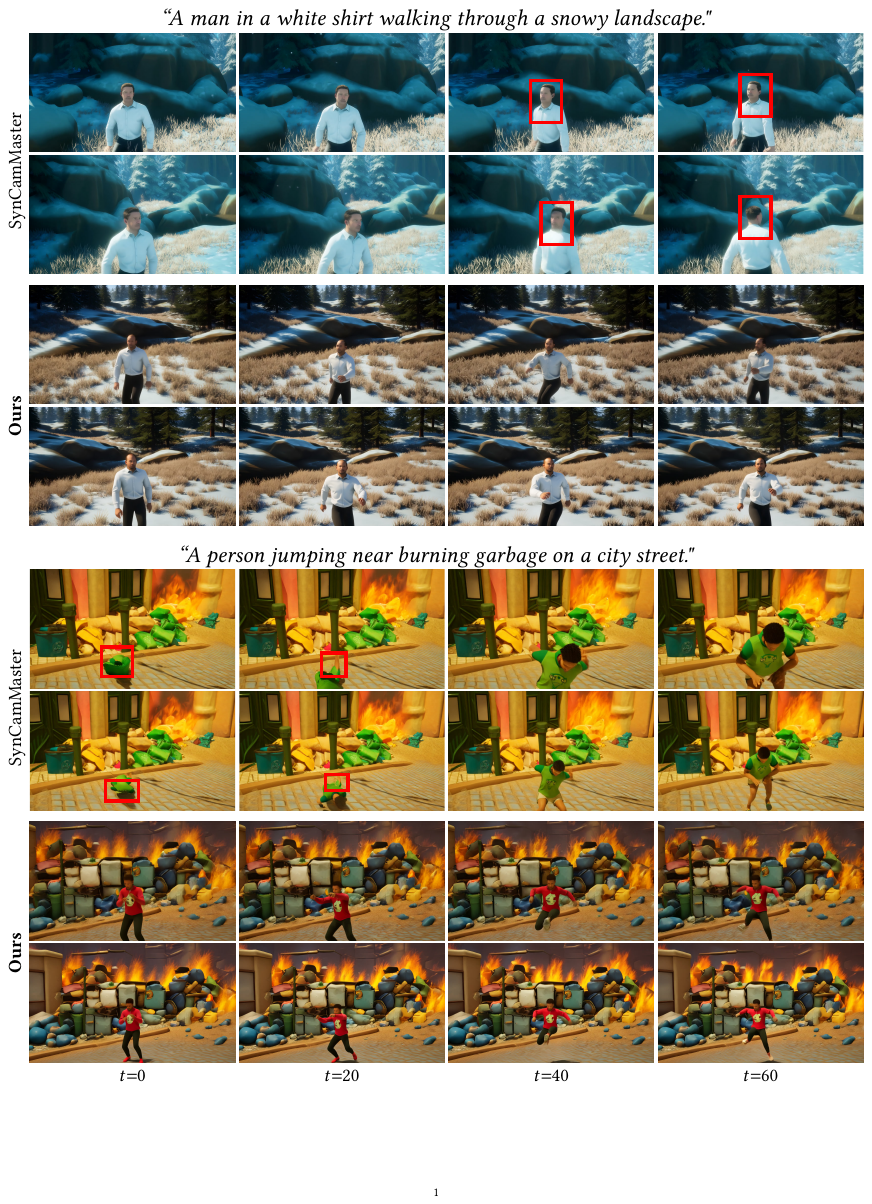}
    \caption{\textbf{Qualitative Comparison on Short Sequences.} We compare MV-Forcing against the bidirectional SynCamMaster~\cite{bai2024syncammaster} teacher at 2 views and 81 frames. For each prompt, we show 2 views (rows) across 4 timesteps (columns). \textcolor{red}{Red} boxes highlight cross-view inconsistencies in SynCamMaster that our method overcomes through the accumulated geometric prior, while achieving comparable visual quality.}
\label{fig:short_comparisons}
\end{figure*}

\smallskip
\noindent \textbf{Short Sequence Comparisons.}
Fig.~\ref{fig:short_comparisons} compares \methodname{} against 
SynCamMaster at 2 views and 81 frames. Our method produces 
comparable visual quality to the bidirectional teacher while 
maintaining stronger cross-view consistency.

\begin{figure*}[t!]
    \centering
    \includegraphics[width=\textwidth, trim={2.7cm 8.5cm 4.0cm 3.2cm}, clip]{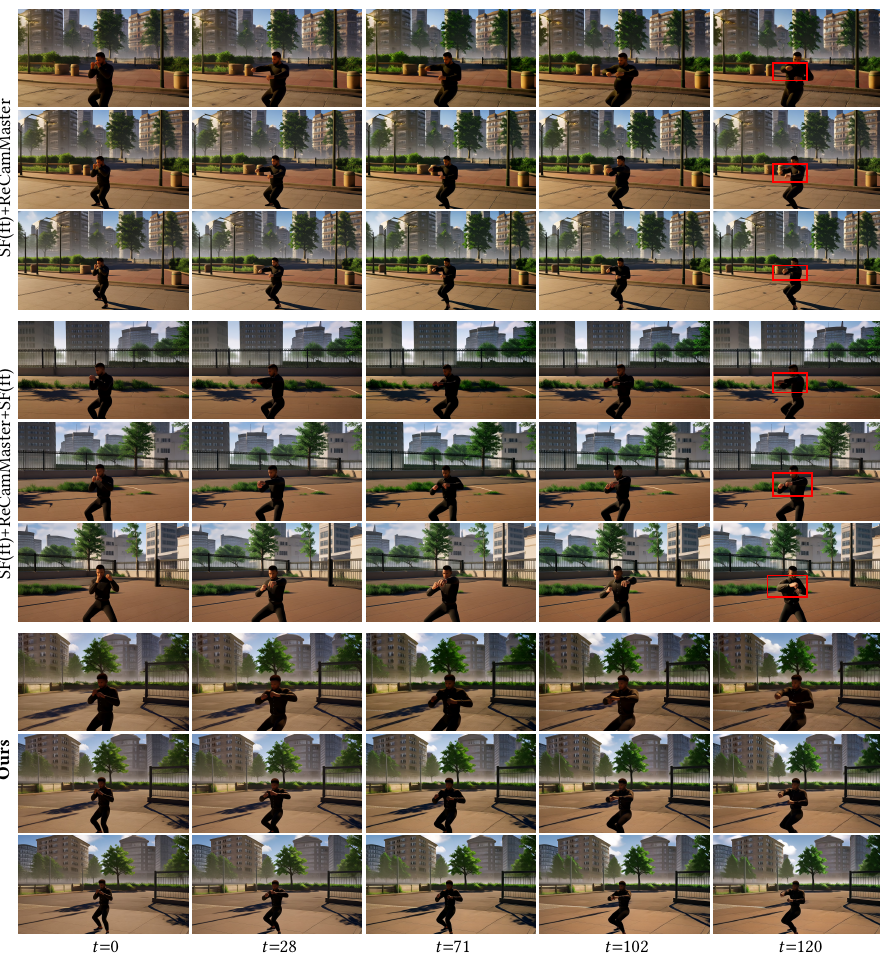}
    \caption{\textbf{Synthetic Long Sequence Comparison.} We generate 3 views across 162 frames for the prompt \emph{``A man practicing martial arts in an urban setting with tall buildings in the background.''} For each method, we show 3 views (rows) at 5 timesteps (columns). All methods are trained and evaluated on the synthetic SynCamVideo dataset for a fair in-domain comparison. SF(ft)+ReCamMaster degrades beyond its 81-frame window, producing visible artifacts in later timesteps. SF(ft)+ReCamMaster+SF(ft) recovers temporal continuity but lacks cross-view consistency. As highlighted by the \textcolor{red}{red} boxes, the baselines produce inconsistent limb positions across views at the same timestep, while our method preserves coherent pose and scene geometry across all viewpoints.}
\label{fig:long_synth_comp}
\end{figure*}

\smallskip
\noindent \textbf{Synthetic Long Sequence Comparisons.}
Fig.~\ref{fig:long_synth_comp} shows long sequence 
comparisons in the synthetic setting at 3 views and 162 frames, 
where we compare against SF(ft)+ReCamMaster and 
SF(ft)+ReCamMaster+SF. The baselines exhibit visible cross-view 
drift and temporal discontinuities, while our method maintains 
consistent appearance and geometry across all views and 
timesteps.

\begin{figure*}[t!]
    \centering
    \includegraphics[width=\textwidth, trim={2.3cm 9.0cm 4.2cm 3.2cm}, clip]{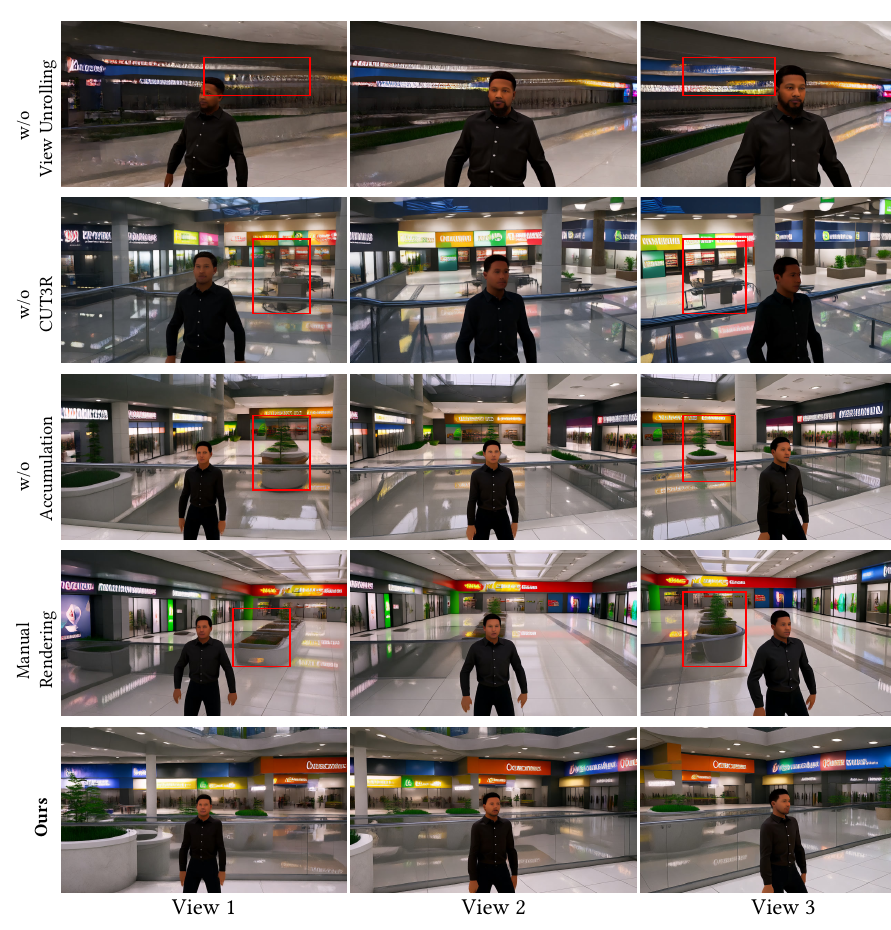}
    \caption{\textbf{Component Ablation.} We show 3 views at a single timestep for each ablation variant described in Sec.~\ref{sec:ablations} in the main paper. \textcolor{red}{Red} boxes highlight cross-view inconsistencies. Without view unrolling, background elements shift unnaturally across views. Without CUT3R, the occluded background behind the person changes between viewpoints. Without accumulation, the tree behind the person shifts forward and changes shape across views. With manual rendering, the tree appears displaced due to warping artifacts in occluded regions. Our full model maintains consistent scene structure across all views.}
    \label{fig:component_ablations}
\end{figure*}

\smallskip
\noindent \textbf{Component Ablations.}
Fig.~\ref{fig:component_ablations} shows qualitative comparisons of 
the ablation variants described in Sec.~\ref{sec:ablations} at 3 
views and 162 frames. Removing view-sequential unrolling and the 
CUT3R geometric prior produce the most visible degradation in 
cross-view consistency.

\begin{figure}[t!]
    \centering
    \includegraphics[width=\textwidth, trim={2.5cm 17.2cm 4.2cm 3.5cm}, clip]{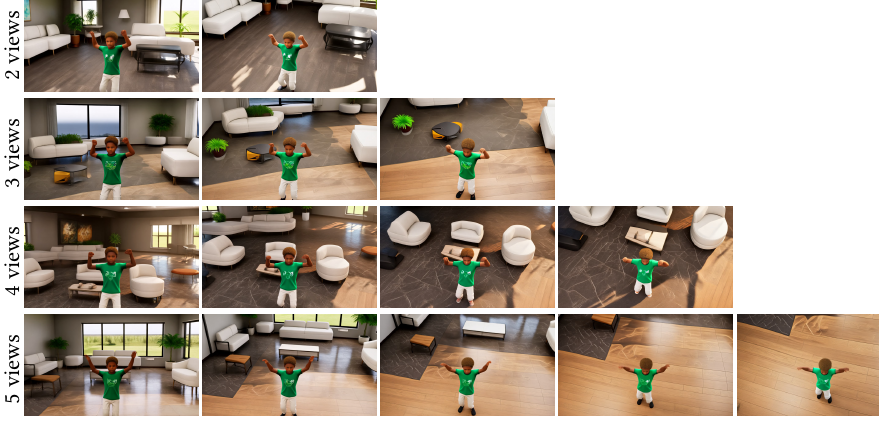}
    \caption{\textbf{View Scaling.} We show all generated views at a single timestep as the number of views increases from 2 to 5 at a fixed 81 frames for the prompt \emph{``A child in a green t-shirt dancing in a modern living room.''} The accumulated geometric prior grounds each new view in the full 4D structure observed so far, maintaining consistent scene geometry as the view count grows.}
    \label{fig:view_scaling}
\end{figure}

\smallskip
\noindent \textbf{View Scaling.}
Fig.~\ref{fig:view_scaling} shows qualitative results as the 
number of views increases from 2 to 5 at a fixed 81 frames. 
Visual quality and geometric consistency remain stable across 
all view counts.

\begin{figure*}[t!]
    \centering
    \includegraphics[width=\textwidth, trim={2.7cm 7.2cm 4.2cm 3.2cm}, clip]{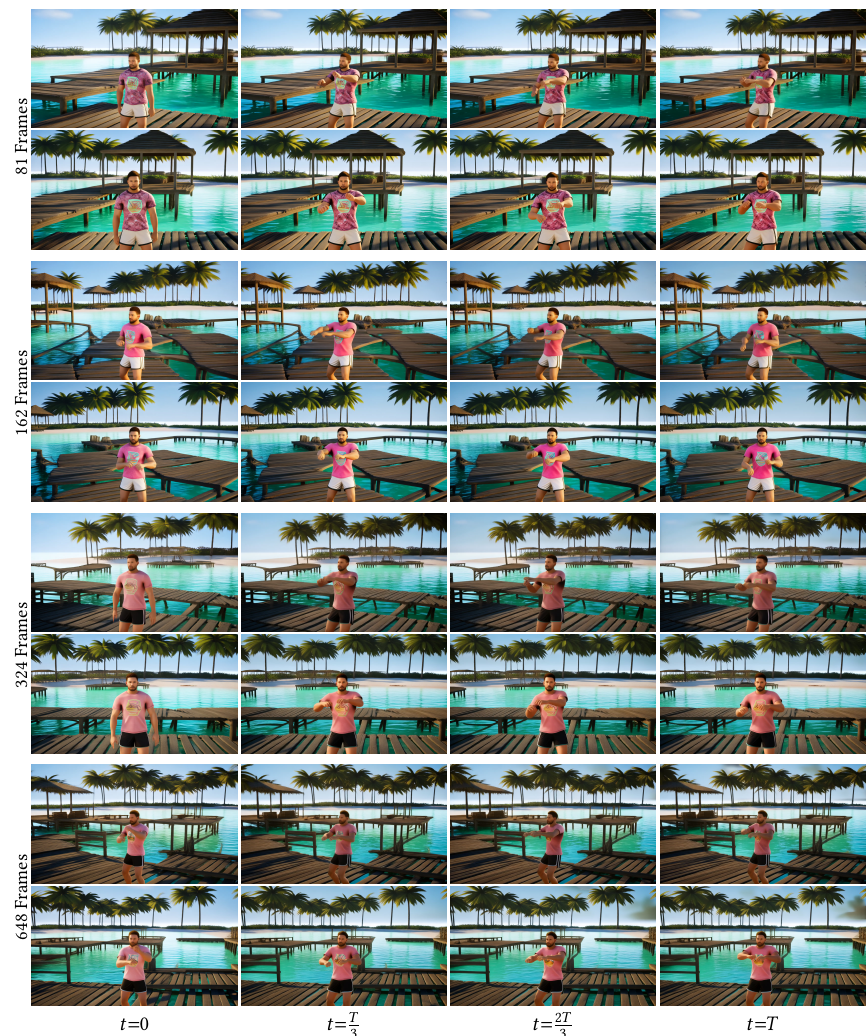}
   \caption{\textbf{Temporal Scaling.} We show 2 views at evenly spaced relative positions as the temporal length increases from 81 to 648 frames for the prompt \emph{``A man in a pink t-shirt practicing martial arts on a wooden deck by the ocean.''}. Cross-view synchronization remains stable across an 8$\times$ increase in duration.}
\label{fig:temporal_scaling}
\end{figure*}

\smallskip
\noindent \textbf{Temporal Scaling.}
Fig.~\ref{fig:temporal_scaling} shows qualitative results as the 
temporal length increases from 81 to 648 frames at a fixed 2 
views. The model maintains consistent appearance and motion over 
an 8$\times$ increase in duration.

\end{document}